\documentclass[sigconf]{acmart}

\newcommand{\tabincell}[2]{\begin{tabular}{@{}#1@{}}#2\end{tabular}}  
\usepackage{natbib}
\usepackage{multirow}
\makeatletter
\def\hlinew#1{%
  \noalign{\ifnum0=`}\fi\hrule \@height #1 \futurelet
   \reserved@a\@xhline}
\makeatother
\AtBeginDocument{%
  \providecommand\BibTeX{{%
    \normalfont B\kern-0.5em{\scshape i\kern-0.25em b}\kern-0.8em\TeX}}}

\begin{document}

\title[MGG-DNN for User Intent Prediction]{TPG-DNN: A Method for User Intent Prediction Based on Total Probability Formula and GRU Loss with Multi-task Learning}

\author{Jingxing Jiang, Zhubin Wang, Fei Fang, Binqiang Zhao}
\affiliation{%
  \institution{Alibaba Group, Beijing $\&$ Hangzhou, China}}
\email{{jingxing.jjx,zhubin.wzb,mingyi.ff,binqiang.zhao}@alibaba-inc.com}

\renewcommand{\shortauthors}{Jingxing Jiang, Zhubin Wang, Fei Fang, Qianyang Zhang, Binqiang Zhao}

\begin{abstract}
The E-commerce platform has become the principal battleground where people search, browse and pay for whatever they want. Critical as is to improve the online shopping experience for customers and merchants, how to find a proper approach for user intent prediction are paid great attention in both industry and academia. In this paper, we propose a novel user intent prediction model, TPG-DNN, to complete the challenging task, which is based on adaptive gated recurrent unit (GRU) loss function with multi-task learning. We creatively use the GRU structure and 
total probability formula as the loss function to model the users' whole online purchase process. Besides, the multi-task weight adjustment mechanism can make the final loss function dynamically adjust the importance between different tasks through data variance. According to the test result of experiments conducted on Taobao daily and promotion data sets, the proposed model performs much better than existing click through rate (CTR) models. At present, the proposed user intent prediction model has been widely used for the coupon allocation, advertisement and recommendation on Taobao platform, which greatly improve the user experience and shopping efficiency, and benefit the gross merchandise volume (GMV) promotion as well. 
\end{abstract}




\maketitle

\section{Introduction}

\begin{figure*}
    \centering
    \includegraphics[width = 13 cm]{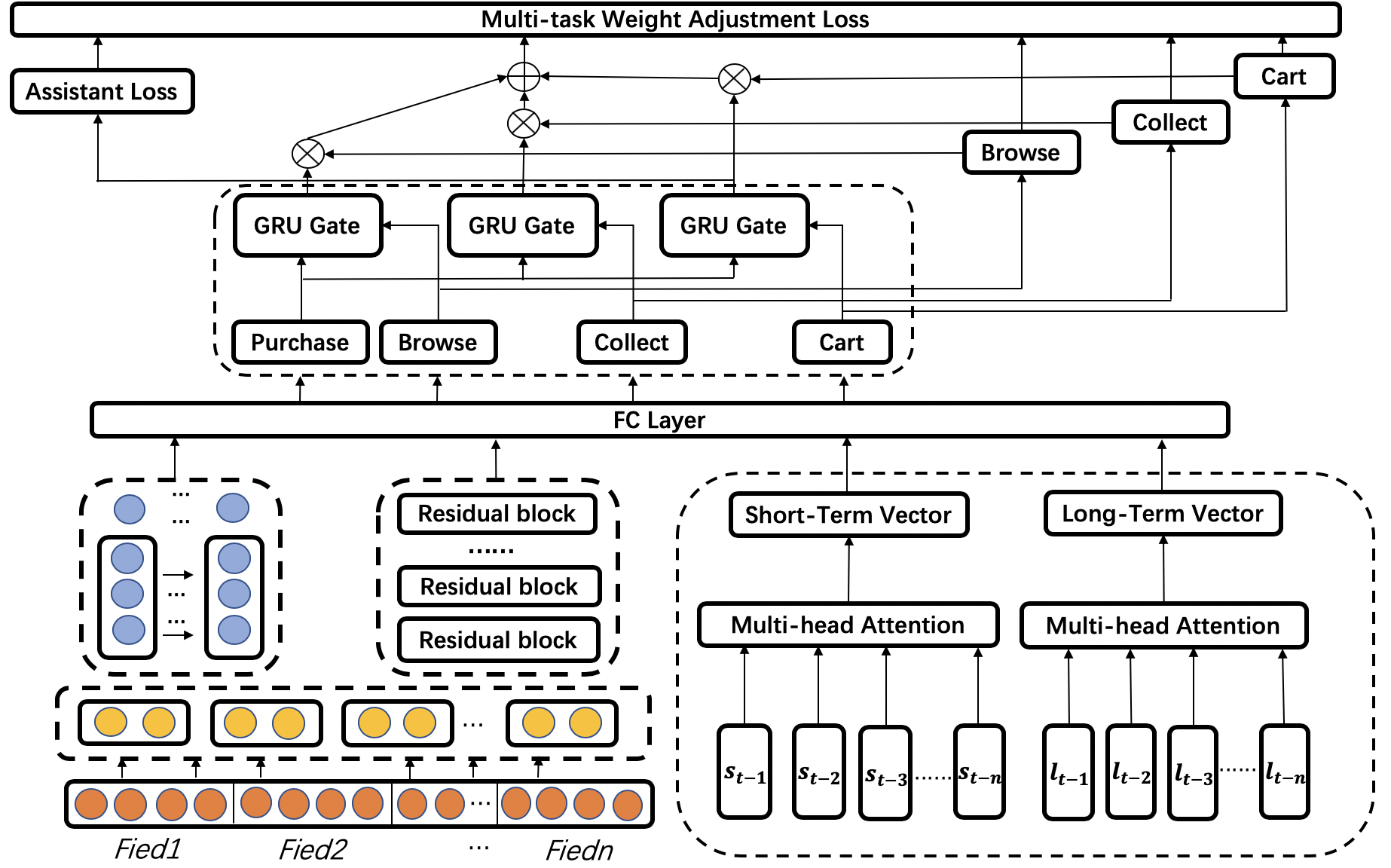}
    \caption{Network architecture}
    \label{fig.archi}
\end{figure*}

Last few years have witnessed a rocket-like increase in e-commerce. As Mainstream large-scale online shopping platforms, Taobao and Amazon are providing great convenience for a huge number of people to purchase products. According to the annual report of 2019 by Alibaba Group \cite{ali}, as of September 2019, Taobao, the largest e-commerce platform in China, has attracted more than $780$ million active users monthly all around the world, with an increase of 104 million over 2018. Highly competitive as e-commerce market evolves,we embrace grand challenges to promote the growing tendency of amount of purchase on platform and meanwhile provide better online shopping experience and efficiency for both customers and merchants. User intent prediction reflects the purchase potential and probability of users based on large amount of user historical behavior data. From the perspective of the platforms, accurate user intent prediction helps to make flexible sales strategy and offer user-specific coupons or product bundles, which surely could inspire and stimulate users to pay for products. While from the perspective of the customers, personalized recommendation significantly improve their purchase experience and efficiency.

We know the purchase behavior is related to {\em browse}, {\em collect} and {\em collect}, a purchase behavior only occurs after them. According to the total probability formula, we can get:
\begin{equation}
\begin{split}
    \boldsymbol{P}(\text{Purchase}) =&\boldsymbol{P}(\text{Browse})*\boldsymbol{P}(\text{Purchase}|\text{Browse})+\\
    &\boldsymbol{P}(\text{Collect})*\boldsymbol{P}(\text{Purchase}|\text{Collect})+\\
    &\boldsymbol{P}(\text{Cart})*\boldsymbol{P}(\text{Purchase}|\text{Cart})
\end{split}
\end{equation}
where $\boldsymbol{P}(\text{Purchase})$ denotes the probability of a user purchase, $\boldsymbol{P}(\text{Purchase}|\text{Browse})$, $\boldsymbol{P}(\text{Purchase}|\text{Collect})$ and  $\boldsymbol{P}(\text{Purchase}|\text{Cart})$
denotes the probability of the conditional probability of the purchase under the corresponding behavior. Since the purchase behavior of users is highly sequential, so we try to fit every conditional probability relationship through GRU structure. On one hand, multi-task learning can not only used for predicting each individual behavior simultaneously, but also help to acquire the association and connection between different sequential parts. On the other hand, better extraction of hidden information from raw data and features achieves better prediction accuracy,which usually depends on the definition of loss function. Deep neural network with a single loss function usually only capture the information needed for individual task while the remaining information hidden in the data is discarded, which is not the way we want, since the user sequential behaviors interplay and interact with each other. As a novel solution, multi-task learning allows different machine learning models to share parameters and perform knowledge transformation according to different loss functions. Hence, multi-task learning can comprehensively capture user behavior information. 

User intent prediction has been well-studied in the literature. \cite{Sebastian2017An} proposed two common patterns of multi-task learning, which are based on hard or soft parameters shared by hidden layers. However, those traditional multi-task learning models only optimize the feature extraction network, regardless of the correlation between different objectives. Afterwards, the entire space multi-task model (ESMM) proposed in \cite{Xiao2018Entire} and the deep Bayesian multi-Target learning (DBMTL) proposed in \cite{Qi2019Deep} began to focus on the correlation between different tasks. Based on the multi-task learning framework, they also build a time series model for  multiple user behaviors. However, they mainly fit the time series correlation with conditional probability (simple Bayesian formula) and full connection neural network. Since the simple Bayesian formula can not extract the related information between different parts of full-link user behaviors, the general fully connected neural network can not model the time-sequential relationship of the behavior. 

The click-through rate (CTR) refers to the probability whether user click to browse when impressed a product or advertise. The post-click conversion rate (CVR) refers to the probability whether user buy after clicking a product or advertise. We build models for both of them called CTR model and CVR model respectively. In order to predict the browse and purchase behavior of users on the whole platform, we introduce the successful CTR and CVR model structures to build the feature information extraction network in our solution. The user behavior information presents a certain periodic regularity. For instance, one's browse, purchase, and collection behaviors in the past few weeks will directly affect his/hers future purchase probability. Therefore, the time series information of the user purchase behaviors plays a significant role in predicting the user intent. By combining the user's basic characteristics and time-series characteristics, we can better estimate the user browse and purchase intent. The general CTR model pays more attention to extracting the hidden cross information between the features, rather than the separate time series features. 

Except from classification tasks such as user purchase and browse, we need to predict regression indicators such as the number of single purchase orders for users. Traditionally, manual regression settings or equal weight distribution methods were often used to separate different regression and classification tasks, in which the loss functions are usually added together. However, dominated by tasks with large gradients during the learning process, those methods are unable to balance all the tasks to be completed. This is also a problem need to deal with in our work.

In this paper, we propose a novel end-to-end total probability formula and adaptive GRU loss function based Deep Neural Network (TPG-DNN) for the user intent prediction. With TPG-DNN, the user behavior features are automatically learned from the raw data without manual operation. We model the purchase probability of user according to the formula of total probability. With multiple loss functions based on the gate recurrent unit (GRU) loss, we could extract the time-series information between user browse, collect, cart behavior and payment behavior. We use heteroscedastic uncertainty to balance the weight between regression loss and classification loss. The contributions of the paper are summarized as follows:
\begin{itemize}
    \item We model the purchase probability of user according to the total probability formula. We conclude that the GRU structure can extract the time-series information and conditional correlation in the process of user purchase prediction. We use GRU loss function to model the conditional probability of the purchase under the corresponding behavior, so that the final multi task loss will help the network fully learn the user time-sequential information.
    \item Based on the principle of maximizing the homoscedastic Gaussian likelihood function, the proposed TPG-DNN can adaptively adjust the weights of different classification tasks and regression tasks in the final loss function, thus the TPG-DNN can fully extract the inherent information of different tasks. Since our final multi task loss function contains two types, regression and classification. This method will reduce the learning deviation caused by inaccurate initial weight setting.
    \item We trained and compared the offline estimation results of the TPG-DNN in the daily periods, and proved the wide applicability of the model. And we not only evaluate the effect of TAG-DNN on large scale offline data, but also apply the model to practical industrial scenarios. The user intent prediction model is used to help Alibaba improve the performance of multiple online scenarios. The results of AB online testing on show that our model is very effective in real industry.
\end{itemize}
The rest of the paper is organized as follows. We show the related work in Section 2. We describe the design of TPG-DNN model in Section 3. We present experiments in Section 4 and followed by the conclusion of this paper in Section 5.

\section{Related Work}
\subsection{Multi-task Learning}
Multi-task learning is widely used in the fields of recommendation systems, natural language processing \cite{16chen2019multi,17collobert2008unified,22collobert2008unified,23deng2013new}, computer vision\cite{18long2017learning}, etc. In \cite{Sebastian2017An}, Ruder proposed two widely used multi-task learning based on hard or soft parameters shared by hidden layers, respectively. In hard parameter based model, a neural network share multiple loss functions under the same network structure. In soft parameter based model, different tasks are trained in different networks. \cite{19sogaard2016deep} find a better multitask structure which the bottom layer of complex tasks should be supervised by low-level task objectives. \cite{20daume2009bayesian} uses Bayes formula to build multi task learning structure. Jiaqi Ma {\em et al.} proposed multi-task learning with multi-gate mixture-of-experts (MMOE) in \cite{MaZYCHC18}. By combined with the widely used Shared-Bottom and MOE structure, MMOE improved the multi-task learning to extract the relevance and difference of different tasks. In \cite{Xiao2018Entire}, multi-task learning is applied to traditional CTR and CVR models to improve prediction accuracy, in which the difficult CVR estimation is transferred into CTCVR estimation and solve the SSB problem in the model. \cite{Qi2019Deep} model conditional probability relationships between different tasks through fully connected neural networks. In the field of image semantic segmentation, \cite{KendallGC18} uses a method of minimizing the variance likelihood function to balance the weight relationship between different tasks.

\subsection{User Intent Prediction Models}
User Intent Prediction Models are widely studied in the industry at present. CTR and CVR models are those of them. They are widely used in the field of computing advertising and recommendation systems to predict user behaviors. CTR and CVR models are mainly used to predict whether users will click and buy on different products and advertisements respectively.
LR and GBDT are the earliest CTR estimation models used in industry. The GBDT + LR model proposed by Xinran He \cite{6he2014practical} combines features through GBDT, and then trained the final model through LR. Rendlep \cite{7steffen2010factorization} proposed a feature cross-information extraction structure for Factorization Machines, which uses the inner product of hidden variables to extract the combined information of features. Afterwards DNN was widely used to fit complex feature interaction structures. \cite{8zhang2016deep} The proposed FNN model extracts the features of the pre-trained FM model, and then sends the crossed features to the DNN. The proposed PNN model in \cite{9qu2016product} mainly adds an inner product layer to the deep learning network to model the correlation between features. These methods focus on extracting high-dimensional information between features, and pay less attention to low-dimensional information of features. Wide and Deep \cite{10cheng2016wide} and DeepFM \cite{11guo2017deepfm} models introduce feature information on the Wide side to enhance the normalization and learning capabilities of the network, allowing the network to learn high- and low-dimensional feature interactions together. \cite{12zhou2018deep} DIN introduces the attention mechanism to learn the weight relationship between users and different products, thereby enhancing the network's ability to fit and express. The DCN network proposed in \cite{13wang2017deep} replaces the original Wide part by using the Cross network. The basic motivation for the design of the Cross network is to increase the interaction between features, and use multiple layers of cross layers to perform feature crossing on the input vector. But the feature interaction granularity of DCN is bit-wise and the compressed interaction network (CIN) structure used in \cite{14lian2018xdeepfm} xDeepFM is vector-wise, which has better feature cross-information extraction capabilities.

\section{Model}
\begin{figure}
    \centering
    \includegraphics[width = 5 cm]{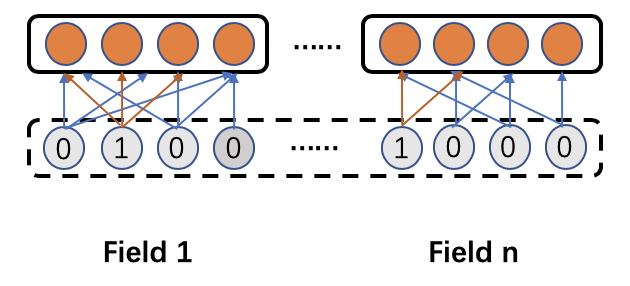}
    \caption{Embedding layer structure}
    \label{fig.embedding}
\end{figure}

We propose a deep multi-task learning network, the total probability formula and Adaptive GRU loss function based Deep Neural Network (TPG-DNN), for the prediction of user behaviors. Fig. \ref{fig.archi} shows the model architecture of TPG-DNN. It begin with an embedding layer which compresses the one-hot vectors of the raw action features into dense vectors. Afterwards, we use three parts to extract feature information. First, a transformer layer is applied to model each time-sequential behavior, which efficiently extracts user time-series information. Second, a CIN layer is applied to model the feature interactions implicitly at the bit-wise level. Third, we use a Deep Residual Network(RESNET) to learn implicit high-order feature interactions. And then, we put all the information to a DNN layer to model the hidden relationship between them. At the same time, a linear regression layer is applied to effectively memorize sparse feature interactions. Finally, TPG-DNN is trained with total probability formula and adaptive GRU-Loss multi-task learning, which use heteroscedastic uncertainty to balance the weights of different tasks in the final loss. We will introduce the details about each part of TAG-DNN as follows.

\subsection{Embedding Layer}
\begin{figure}
    \centering
    \includegraphics[width = 5 cm]{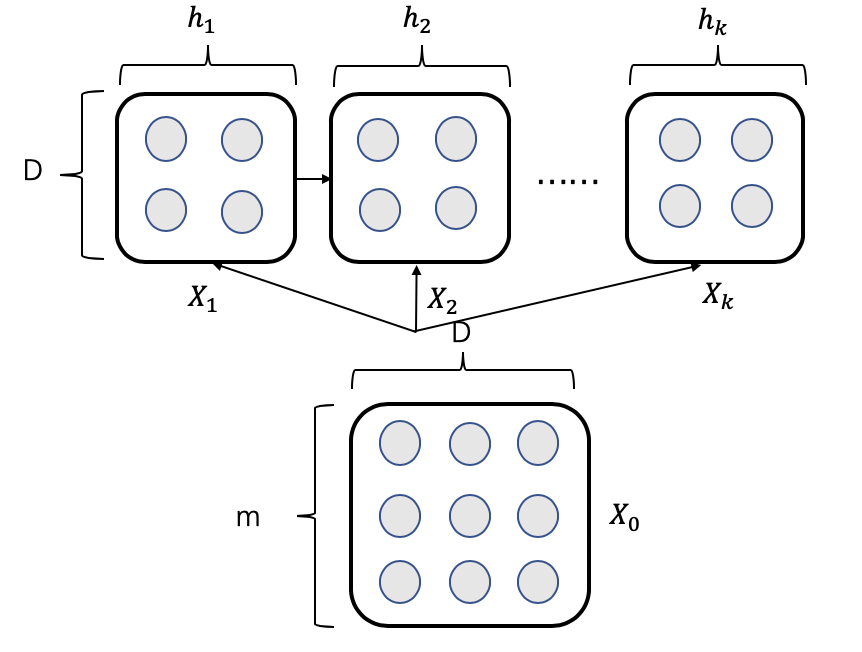}
    \caption{CIN structure}
    \label{fig.cin}
\end{figure}
For deep learning, data cleaning and feature engineering are the most important step. Through statistical analysis of a large amount of data, we found that several field features play fundamental role in the prediction of user purchase probability, mainly including: user basic features, user daily behavior features, user monthly behavior features, user grade behavior features.

Our model incorporates features from the full spectrum of customer
information available at Taobao. They are made up of four parts from the data,including: (1) users' basic portrait information, (2) users' behavior history, (3) users' historical behavior in the promotion over the years

In users' basic portrait information, we use the user's basic information, including gender, age group, purchase level in the past year, province of residence and so on.

In users' behavior history, the user's behavior features directly describe the user's activity in the situation. We focus on what features are used to describe the user's activity, including users' purchase behavior in the past few time, the number of days since registration, the maximum number of consecutive active days, as well as the changes in the activity in the past few time. The changes are used to describe the variation trend of the user's activity. 

In users' historical behavior in the promotion, considering that people who buy in previous sales promotion may have stronger purchase intention in the future. So we add some features which describes their behavior purchases in the past sales promotion to enrich characteristic system.

We divide the numerical variables into boxes and convert all features into one-hot vectors, which will be used as the input of all feature extraction layers. Because of the large number of features, a single one-hot vector will cause the data to be sparse, so we use embedding operation to reduce the dimension of each field feature.

The embedding layer structure is shown in Fig. \ref{fig.embedding}. Features from different $M$ fields are compressed by the embedding layer into vectors, $\boldsymbol{e}_i\in\mathbb{R}^{n_i} \forall\;\; i = 1\cdots M$, where $n_i$ is the number of output dimension for the feature from each field. Stack all outputs, we have $\boldsymbol{E}_m = [\boldsymbol{e}_1,\boldsymbol{e}_2,\cdots\boldsymbol{e}_M]^T$. We use $\boldsymbol{E}$ as the input of the followed feature information extraction layer and it is trained within the model.

\subsection{Feature Information Extraction Layer}
\textbf{Cross information extraction component.} Feature crossing is a common feature information learning structure in deep learning. Through the powerful computing and fitting ability of deep learning, we can learn the cross combination between different features automatically for the final prediction target.

At present, DCN is often used to extract feature crossing information. It can learn the cross relationship between features without too much pretreatment. But it is bit-wise. For example, the embedding vector of the city field is < A1, B1, C1 >, the embedding vector of the age field is < A2, B2, C2 >. In the cross layer, A1, B1, C1, A2, B2, C2 will be directly used as input after connecting. So it does not realize the concept of field vector. While CIN\cite{14lian2018xdeepfm} has two special virtues comapred with DCN: (1) it can learn certain bounded-degree feature interactions, (2) it learns feature interactions at a vector-wise level. So we use the CIN structure to extract cross features. 

The input of the CIN layer comes from the embedding layer. Assuming that there are many fields, and the embedding vector dimension of each field is D. The structure of CIN is shown in Fig. \ref{fig.cin}. Denote $\boldsymbol{X}_{h,*}^{k}$ as the output of the $k$-th layer, we have:
\begin{equation}
    \boldsymbol{X}_{h,*}^{k} = \sum_{i=1}^{H_{k-1}}\sum_{j = 1}^{m}\boldsymbol{W}_{i,j}^{k,h}(\boldsymbol{X}^{k-1}_{i,*}\circ \boldsymbol{X}^{0}_{j,*})\in\mathbb{R}^{1\times D}\;\;\forall 1\cdots H_k ,
\end{equation}{}
where ${W}_{i,j}^{k,h}$ is the weights and $\circ$ represent the Hadamard product between two vectors. The CIN structure is a vector-wise weighted procedure after the hadamard product between $\boldsymbol{X}^{k-1}_{i,*}$ and $\boldsymbol{X}^{0}_{j,*}$.

\begin{figure}
    \centering
    \includegraphics[width = 5.5 cm]{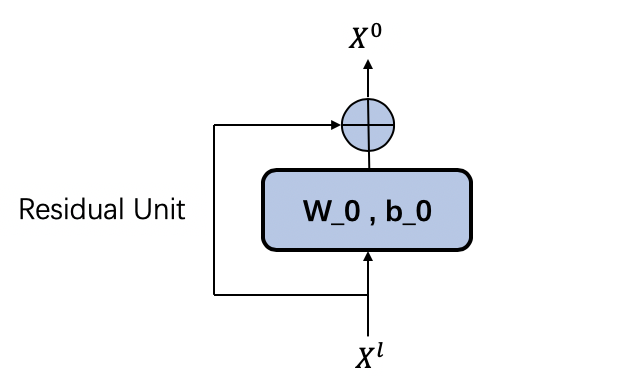}
    \caption{A basic residual block of RESNET}
    \label{fig.deep}
\end{figure}
\textbf{Deep and Linear component.} In deep network, fully connected neural network is used to learn high order feature interactions. However deeper network will also introduce the problems such as parameter explosion, gradient disappearance, and even over fitting. RESNET has been widely used in image recognition. It use residual block to solve the problem of network degeneration, which may be of great help for us. Therefore, we improve the deep side, through the residual connection unit in RESNET to enhance the fitting ability of the deep network.

The input of the Deep and Linear component is the vector processed by Embedding component as shown in Fig. \ref{fig.deep}. It is a basic residual block of RESNET, i.e., $X^0 = f(X^l)+X^l$. If those layers behind the deep network are identity maps, it is difficult for nonlinear activation function to fit identity map. While in residual block, we just need to make the map equal to zero. The residual connection structure can make the network composition deeper, so that the network can learn more characteristic information than the general full connection network. 

The linear part uses the traditional linear regression structure, i.e., $Y = WX + b$. In the selection of activation function, we use parametric relu [15]. It adjusts the activation range of Relu in an adaptive way. The structure is as follows:
\begin{equation}
     \text{Relu}=\left\{
\begin{aligned}
x & ~~\text{if}~x>0 \\
\alpha x & ~~\text{if}~x\leq0
\end{aligned}.
\right.
\end{equation}

As a random super parameter, $\alpha$ can be learned through network. Parametric Relu can be transformed into Relu or leaky Relu according to the actual data.

\begin{figure}
    \centering
    \includegraphics[width = 6 cm]{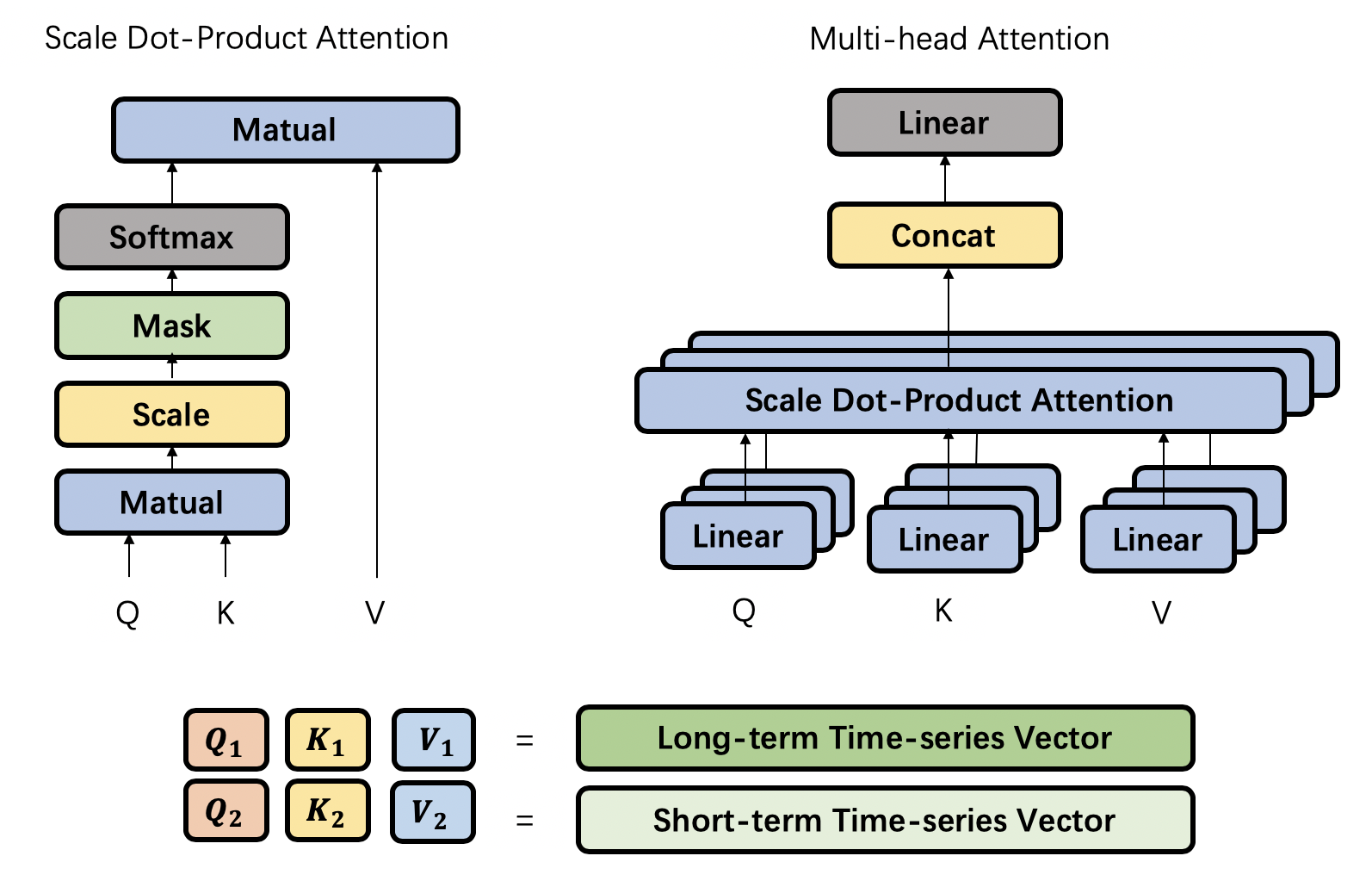}
    \caption{Transformer structure}
    \label{fig.trans}
\end{figure}
\textbf{Transformer component.} From the long-term analysis and observation, we find that the purchase behavior of users has a strong time regularity. Therefore, we will extract the time-series information of users separately through the Transformer structure. We will divide the time-series information of users into two parts, one is short-term timing rule, and the other is long-term timing rule.

The short-term time-series information of each user can be represented as $S = \{S_1, S_2, \cdots S_t\} \in \mathbb{R}^{t\times d}$, where $t$ is the number of recorded days and $d$ is the number of behaviors, including purchase, browse and adding collection. Short-term time-series information describes the user behavior during short period (a few days). The long-term time-series information of each user can be denoted as $S = \{L_1, L_2, \cdots L_T\} \in \mathbb{R}^{T\times d}$, where $T$ represents the period and $d$ is the number of behaviors. Long time-series features show the user behavior in during a long time. 

The transformer model \cite{21vaswani2017attention} is currently widely used in the field of natural language processing, especially in machine translation problems. As shown in the Fig. \ref{fig.trans}, a transformer model is essentially a multi-headed self-attention structure connected by multiple expressions. The attention structure in it is
\begin{equation}
    \text{attention}_{\text{output}} = \text{Attention}(Q,K,V) = \text{Softmax}(\frac{QK^T}{\sqrt{d_k}})V.
\end{equation}

In natural language process, Q means Query. K means Key. V means value. The multi-head attention is the connection of different attention results, which is the linear tranformation of $Q$, $K$ and $V$. In particular,
\begin{equation}
    \text{MultiHead}(Q,K,V) = \text{Connect}(\text{head}_1,\text{head}_2,\cdots,\text{head}_i)W,
\end{equation}
and 
\begin{equation}
    \text{head}_i = \text{Attention}(QW_i^Q,K_i^K,V_i^V).
\end{equation}

Time-series information is similar to natural language, which is composed of sequence information with context. Therefore, we apply the multi-headed self-attention structure to the extraction of time-series information and $Q$, $K$ and $V$ share the same time-series vector. The short-term time-series feature vectors and long-term time-series feature vectors are respectively input into two independent transformer structures for embedding, so as to obtain new long-term and short-term time series feature vectors $S^*$ and $L^*$. We perform max pooling on two sets of time-series feature sets to complete the time-series information extraction.

\subsection{Multi-task Component}
\begin{figure}
    \centering
    \includegraphics[width = 6 cm]{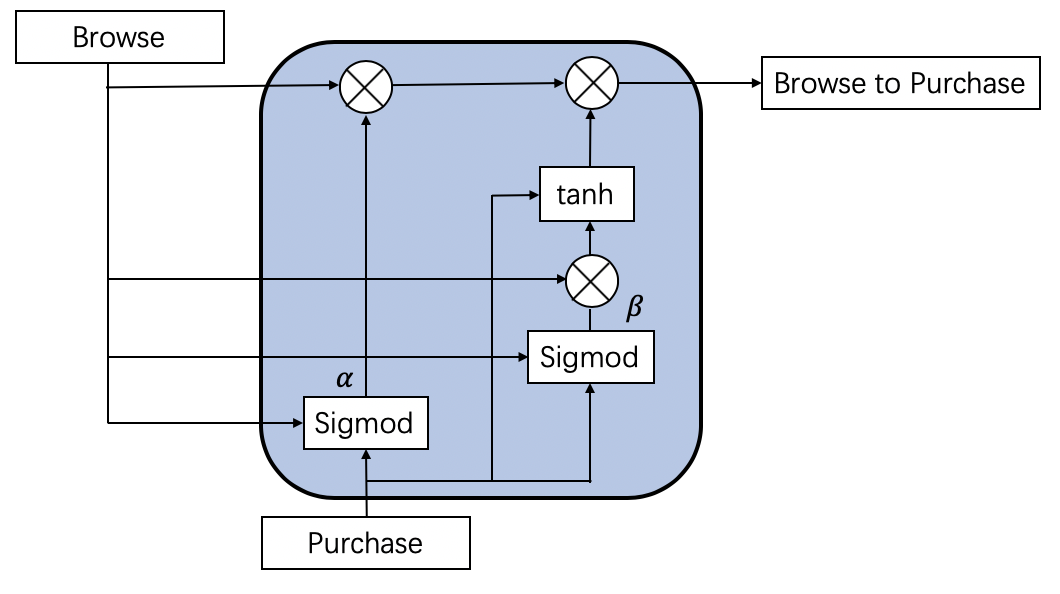}
    \caption{BBPBR probability fit model}
    \label{fig.BBPBR}
\end{figure}
After all the information passes through the feature information extraction layer, three groups of different information vectors will be obtained, including: the feature cross information vector, the feature time-sequential information vector and the feature linear information vector. We will gather the first two types of information into a fully connected neural network, so as to learn the interactive information between different high-dimensional feature information vectors. Then the out put of the fully connected neural network and the linear feature information vector are summed. So the high-dimensional and low-dimensional information extracted from the network are fused to complete the feature information extraction. 

Multi task learning is a kind of transfer learning. Its basic definition is: the model loss is consisted of M learning tasks, all or part of them are related but not exactly the same. The goal of multi task learning is to help improve the performance of each task by using the knowledge contained in the M tasks. We build a multi task learning layer through four interrelated basic tasks. It includes three classification tasks and one regression task.

In the process of online shopping, customers' feedback can be divided into three categories:Browse Behavior (BBR), Collect Behavior (CBR), Cart Behavior (CAR), Purchase Behavior (PBR). Loss function directly determines the learning direction and goal as an optimization objective. A comprehensive fusion of distinct goals to the final loss function can make the model fully learn the user's behavior pattern, thus getting better prediction effect. Therefore, for the multi task learning framework, we propose three classification optimization objectives, namely BBR, CBR, CAR and PBR. At the same time, in the user's purchase behavior, the volume of each purchase plays a critical role in depicting the user's value. So we also add order volume to our loss function, which is a regression task.

The objective function for the Browse Behavior (BBR) optimization is
\begin{equation}
    L_{\text{BBR}} = -\frac{1}{N}\sum_{j=1}^{N}\log(\boldsymbol{P}(b_j|X,\theta)).
\end{equation}

The objective function for the Collect Behavior (CBR) optimization is
\begin{equation}
    L_{\text{CBR}} = -\frac{1}{N}\sum_{j=1}^{N}\log(\boldsymbol{P}(c_j|X,\theta)).
\end{equation}

The objective function for the Cart Behavior (CAR) optimization is
\begin{equation}
    L_{\text{CBR}} = -\frac{1}{N}\sum_{j=1}^{N}\log(\boldsymbol{P}(a_j|X,\theta)).
\end{equation}

The objective function for the Purchase Behavior (PBR) optimization is
\begin{equation}
    L_{\text{PBR}} = -\frac{1}{N}\sum_{j=1}^{N}\log(\boldsymbol{P}(u_j|X,\theta)).
\end{equation}

$\boldsymbol{b}_{j}$ is random variable which stands for browse behavior. $\boldsymbol{c}_{j}$ is random variable which stands for collect behavior. $\boldsymbol{a}_{j}$ is random variable which stands for cart behavior. ${u}_{j}$ is random variable which stands for purchase behavior. $\boldsymbol{X}$ means features matrix. $\boldsymbol\theta$ means parameter.

The objective function for the order volume optimization is
\begin{equation}
    L_{\text{OV}} = \frac{1}{N}\sum_{j=1}^{N}(O_i - f(x_i))^2,
\end{equation}
where ${Q}_{i}$ stands for the order volume, ${X}_{i}$ means the i-th row of features matrix and $f(x)$ stands for the output of the model.

\textbf{Total probability GRU-Loss component.} We know that the users' purchase behavior is directly affected by the other three related behaviors. The users' browsing, purchasing and collecting behaviors all play significant conditional probability relationship in the user's final purchase intention. And the influence of these three behaviors on purchase behavior can be considered as equal to each other. We can express this relationship through total probability formula such an expression:
\begin{equation}
\begin{split}
    \boldsymbol{P}(\text{Purchase})=\boldsymbol{P}(u_j|X,\theta)=&\boldsymbol{P}(\text{Browse})*\boldsymbol{P}(\text{Purchase}|\text{Browse})+\\
    &\boldsymbol{P}(\text{Collect})*\boldsymbol{P}(\text{Purchase}|\text{Collect})+\\
    &\boldsymbol{P}(\text{Cart})*\boldsymbol{P}(\text{Purchase}|\text{Cart})
\end{split}
\end{equation}
where $\boldsymbol{P}(\text{Purchase})$ denotes the probability of a user purchase, $\boldsymbol{P}(\text{Purchase}|\text{Browse})$, $\boldsymbol{P}(\text{Purchase}|\text{Collect})$ and  $\boldsymbol{P}(\text{Purchase}|\text{Cart})$
denotes the probability of the conditional probability of the purchase under the corresponding behavior. For the conditional probability, we try to use GRU loss function to model them. Compared with the general multi task loss function, we objectively consider the influence of related tasks on the main task.

As we know, GRU and Long short-term memory (LSTM) are widely used in the modeling process of time series information due to their special gate structure mechanism. The core of gate structure is the control of information. Different gates leave meaningful information in the previous state through deep network. In our scenario, we think that the user's collecting, carting and browsing behaviors are directly related to the user's purchase behavior, but not all the information of collecting, carting and browsing has a positive meaning for purchase. In many cases, the behavior of users' collecting, carting and browsing is just because of some very random reasons, such as operation errors. In this case, a general conditional probability formula cannot accurately model this part of information. So we try to filter the condition information that is valuable to purchase behavior through the gate structure. The network can filter the condition information that is valuable to purchase behavior by itself. Because our training data is about 500 million and features 
are more than 500 dimensions. So we use a more efficient GRU instead of LSTM. 

For the probability of the conditional probability of the purchase under collect task, we can try to use GRU to fit its probability. Browse stands for collection task output through network learning. Purchase stands for purchase task output through network learning. This structure can not only model the impact of conditional probability between browse and purchase behaviors, which makes full use of the sequence information between the two behaviors, but also help the latter stage make the most of the detailed information of the previous stage. The probability fitting structure of Purchase-Browse is shown in Fig. \ref{fig.BBPBR}, in which we define
\begin{equation}
\begin{split}
    &H_{\text{Purchase-Browse}} = \alpha*H_{\text{Browse}} + (1 - \alpha) * H^{'}_{Purchase}\\
    &\alpha = \sigma(W_{\alpha}[H_{\text{Browse}}, H_{\text{Purchase}}]),
\end{split}
\end{equation}
and
\begin{equation}
\begin{split}
    &H_{\text{Purchase}}^{'} = \tanh(W\dot[\gamma*H_{\text{Browse}},H_{\text{Purchase}}])\\
    &\gamma=\sigma(W_{\gamma}[H_{\text{Browse}}, H_{\text{Purchase}}]).
\end{split}
\end{equation}

As a result, the probabilities of the user intent can be given as
\begin{equation}
    \boldsymbol{P}(\text{Purchase}|\text{Browse}) = \sigma(H_{\text{Purchase-Browse}})
\end{equation}
where $\sigma(\cdot)$ is the sigmod function. $\alpha$ and $\gamma$ are defined by both Browse task specific features and Purchase task specific features. We can also get the conditional probability between collect,cart and purchase behaviors:
\begin{equation}
    \boldsymbol{P}(\text{Purchase}|\text{Collect}) = \sigma(H_{\text{Purchase-Collect}})
\end{equation}
\begin{equation}
    \boldsymbol{P}(\text{Purchase}|\text{Cart}) = \sigma(H_{\text{Purchase-Cart}})
\end{equation}

In the multi-task learning model, the weights of different loss functions are also parameters that needs to be adjusted. In the previous literature, experience is often used to determine the weight of different losses in the total loss function. Such an approach may prevent the total loss function from accurately positioning the importance of different problems. We use a weight adjustment mechanism of the maximum likelihood heteroscedastic uncertainty\cite{KendallGC18} to determine the total loss function,which has been used in the field of image semantic segmentation.

For the regression task, we define the Gaussian likelihood function as
\begin{equation}
    \boldsymbol{P}(y|f(x))= N(f(x),\sigma),
\end{equation}
and the logarithm Gaussian likelihood function as
\begin{equation}
    \log(\boldsymbol{P}(y|f(x)))= -\frac{1}{\sigma^2}(y-f(x))^2-\log(\sigma).
\end{equation}

For classification tasks, we use the softmax function to normalize the output of the model to get the probability distribution,
\begin{equation}
    \boldsymbol{P}(y|f(x)) = \text{Softmax}(\frac{1}{\sigma^2}f(x)),
\end{equation}
and the logarithm function is
\begin{equation}
    \log(\boldsymbol{P}(y|f(x))) = \frac{1}{\sigma^2}f_k(x) - \log(\sum_{k^{'}}\exp{(\frac{1}{\sigma}f_{k^{'}}(x)))},
\end{equation}
where $\sigma^2$ is a normalization factor, which shows the uncertainty of the data. Rewrite objective functino into 
\begin{equation}
\begin{split}
    L & = -\log(\boldsymbol{P}(y_{\text{BBR}},y_{\text{PBR}},y_{\text{BBPBR}},y_{\text{OV}}|f(x)))\\
    & = -\log\prod_{i = 1}^{3}\text{Softmax}(y_i = k_i;f(x),\sigma^2)*N(y_{OV};f(x),\sigma_4^2)\\
    & = -\sum_{i=1}^{3}\log(\boldsymbol{P}(y_i = k_i|f(x),\sigma^2)) + \frac{1}{2\sigma^2_{4}}(y_{OV}-f(x))^2 + log(\sigma_4)\\
    & \approx \sum_{i=1}^{3}\frac{1}{\sigma_i^2}L_{i}(W)+\frac{1}{2\sigma_4^2}L_4(W)+\log(\sigma_1\sigma_2\sigma_3\sigma_4),
\end{split}
\end{equation}
where $L_i(W) = \text{Softmax}(y_i;f(x))$ represents the loss function of cross entropy and $L_4(W) = (y_{OV}-f(x))^2$ represents the loss function of regression task. The weigh of each task is decided by its variance.

\section{Experiment}
In this section, we present a comprehensive evaluation of the performance for TPG-DNN. We first introduce the experimental setup and then present the experimental results.
\subsection{Experimental Setup}
\textbf {Dataset Statistics.} Our training set and test set are all from real Taobao historical transaction data. These data sets contain the real search , browse, collect, purchase and other user behaviors on Taobao every day. 

In daily user behavior estimation, we use data at t-1 day as training data. Data at t+1 day is used as testing data. User behaviors occurred during the annual sales promotions period have a strong correlation and regularity compared with that in the last year. Training data and testing data are randomly sampled. They both contain 200 million user behavior data.

\textbf{Compared methods.} We compare TPG-DNN to the state-of-the-art approaches in purchasing intent prediction. In the following, we introduce the compared methods briefly.
\begin{itemize}
    \item \textbf{LR}: The traditional linear regression model which takes all the features as input, and use the purchase behavior as labels.
    \item \textbf{xDeepFM}: A Network structure of cross feature extraction by CIN structure
    \item \textbf{MTL}: The basic multi task learning model which excludes time-series feature information extraction layer and adaptive GRU multi-task loss function. The regression loss of order volume was compared with MSE.
    \item \textbf{ESMM}: A multi task model which is widely used in the CTR and CVR prediction. The task of ctcvr is estimated by Bayesian formula, so as to avoid SSB problem caused by direct prediction of CVR task. 
    \item \textbf{MMOE}: A multi task learning structure for information extraction through multi gate control mechanism.
    \item \textbf{TPG-DNN}: The new multi task learning model which is proposed in this paper. Time series information extraction layer is added to feature extraction layer, and total probability GRU loss function is added to the final loss function. The final loss function is obtained by weighted fusion of different losses through the uncertainty mechanism.
\end{itemize}

\textbf{Evaluating Indicator.} For classification tasks, we use AUC and F1 as evaluation indicators for model effects. For regression tasks, we use MAE, MAPE, and WMAPE as evaluation indicators. In particular, we have
\begin{equation}
    \text{MAE} = \frac{1}{n}\sum^{n}_{i = 1}|y_i^{'}-y_i|,
\end{equation}
\begin{equation}
    \text{MAPE} = \frac{1}{n}\sum^{n}_{i = 1}\frac{|y_i^{'}-y_i|}{y_i},
\end{equation}
and
\begin{equation}
    \text{WMAPE} = \frac{\sum^{n}_{i = 1}|y_i^{'}-y_i|}{\sum^{n}_{i = 1}y_i}.
\end{equation}

For MAPE, it has two disadvantages. First, we found that if the divisor is 0, this indicator can not be calculated. Second, because of the difference in the order of magnitude, the influence of the same error on the final result is different in the order of magnitude. Therefore, we also introduced WMAPE to evaluate the regression effect. Since it is a weighted measure, it does not have the same problem as MAPE. For MAPE, if the order quantity is 0, the denominator of MAPE is 1.

\subsection{Experimental Results}
\begin{table*}[]
    \centering
    \begin{tabular}{c|c|c|c|c|c|c|c|c}
    \hlinew{1 pt}
    \multirow{2}{*}{Method}&
    \multicolumn{2}{c|}{Purchase}&\multicolumn{2}{c|}{Browse}&\multicolumn{2}{c|}{Cart}&\multicolumn{2}{c}{Collect}\cr\cline{2-9}
    &AUC&F1&AUC&F1&AUC&F1&AUC&F1\cr
    \hline
    \hline
    LR&0.75774&0.47812&0.78071&0.88662&0.78222&0.29681&0.83219&0.23369 \cr\hline
    xDeepFM&0.77325&0.47263&0.78115&0.88257&0.79261&0.30481&0.85516&0.24993 \cr\hline
    MTL&0.77983&0.49623&0.78581&0.88607&0.79382&0.30519&0.85571&0.24813\cr\hline
    ESMM&0.78032&0.49812&0.78612&0.88637&0.79403&0.30642&0.85592&0.24876 \cr\hline
    MMOE&0.78135&0.49971&0.78672&0.88927&0.79513&0.30781&0.85612&0.25878 \cr\hline
    \textbf{TPG-DNN}&\textbf{0.79128}&\textbf{0.51229}&\textbf{0.79082}&\textbf{0.88973}&\textbf{0.79583}&\textbf{0.30891}&\textbf{0.85618}&\textbf{0.25963} \cr\hlinew{1 pt}
    \end{tabular}
    \caption{AUC and F1 comparison of purchase,browse,cart and collect}
    \label{tab.1}
\end{table*}

\begin{table}[]
    \centering
    \begin{tabular}{c|c|c|c}
    \hlinew{1 pt}
    \multirow{2}{*}{Method}&
    \multicolumn{3}{c}{Order Volume}\cr\cline{2-4}
    &MAE&MAPE&WMAPE\cr
    \hline
    \hline
    LR&0.24331&0.45211&0.48368\cr\hline
    xDeepFM&0.23872&0.39859&0.38774\cr\hline
    MTL&0.15529&0.19275&0.21883\cr\hline
    ESMM&0.12576&0.18255&0.21522\cr\hline
    MMOE&0.13772&0.15618&0.17743\cr\hline
    \textbf{TPG-DNN}&\textbf{0.11407}&\textbf{0.13711}&\textbf{0.14281}\cr\hlinew{1 pt}
    \end{tabular}
    \caption{Order Volume comparison in daily period}
    \label{tab.2}
\end{table}

Table \ref{tab.1} and Table \ref{tab.2} shows the effect of different models on daily user data. Several points can be seen: (1) TPG-DNN outperforms the traditional LR models in the prediction effect of user browse, collect, cart, purchase and order quantity. In purchase behavior estimation, AUC increases by \textbf{3.354$\%$}. In browsing behavior estimation, AUC increased by \textbf{1.011$\%$}. In carting behavior estimation, AUC increased by \textbf{1.361$\%$}. In terms of user order volume estimation, MAPE decreased by 0.129. WMAPE decreased by 0.341. Because our data scale is 200 million, so this improvement is very significant. At the same time, it also shows that the new multi-task prediction framework can better describe user behavior than traditional modle.(2) TPG-DNN is better than MTL, ESMM, MMOE in the prediction effect of user intent behavior. In purchase behavior estimation, AUC is increased from \textbf{1.145$\%$} to \textbf{0.993$\%$} respectively. It can be seen that our total probability GRU loss can more fully learn the correlation between user behaviors than ESMM and MMOE.

In summary, time series feature extraction network, total probability GRU loss function and variance uncertainty weight adjustment mechanism have positive effects on the prediction of user behavior estimation. With the increase of these modules, the effect of the network gradually improves.

\subsection{Online Results}
\begin{table}[]
    \centering
    \begin{tabular}{c|c|c|c|c}
    \hlinew{1 pt}
    Task & \tabincell{c}{Single\\Cost} & \tabincell{c}{Verification\\Rate } & ROI & \tabincell{c}{Order\\Volume}\cr
    \hline
    \hline
    \tabincell{c}{Daily Red \\Pockets Allocation}& -3.04\% & - & +24.33\% & +16.62\% \cr\hline
    \tabincell{c}{Daily Coupon\\Allocation} & - & +83.40\% & - & +4.55\% \cr\hline
    \tabincell{c}{Double 9\\Promotion} & - & - & +241\% & -\cr\hline
    \tabincell{c}{Double 11\\Promotion}& - & +10.44\% & - & -
    \cr\hlinew{1 pt}
    \end{tabular}
    \caption{Online AB testing results in daily or promotion period}
    \label{tab.5}
\end{table}

Coupon allocation is a very significant promotion method in e-commerce platform. In this subsection, we will introduce a strategy based on the user intent prediction by TPG-DNN in online traffic of Taobao. We also introduce the evaluation from daily and promotion period. 

We set two coupon allocation strategies to compare the performance, defined as follows:
\begin{itemize}
    \item \textbf{Random Strategy}: Everyone in this bucket get coupon randomly. If we provide red pocket, the amount of the red pocket is randomly. 
    \item \textbf{Model Strategy}: Everyone in this bucket get coupon by TPG-DNN. We use the score predicted by TPG-DNN as thresholds to decide the allocation.
\end{itemize}

Verification rate ($V_{r}$), cost per order ($C_{o}$), return of investment (ROI) are used as evaluation metrics, defined as follows:
\begin{equation}
    {V_r} = \frac{V_e}{R_e}.
\end{equation}
where $V_{e}$ stands for verification which means the number of coupon people used. $R_{e}$ stands for reception which means the number of coupon people received. 

\begin{equation}
    {C_o} = \frac{V_r}{O_v}.
\end{equation}
where $V_{r}$ stands for verification from red pocket which means the number of money people used by red pocket. $O_{v}$ stands for order volume.

\begin{equation}
    \text{ROI} = \frac{T_r}{V_r}.
\end{equation}
where $T_{r}$ stands for transaction led by red pocket.

\textbf{Daily Period.} We used TPG-DNN in two main scenarios
, including red pockets allocation and coupon allocation. 

In daily red pockets allocation, the model strategy can effectively improve the $C_{o}$, $O_{v}$ and ROI. Model strategy decreased $C_{o}$ by \textbf{3.04$\%$} compared by random strategy. Model strategy increased $O_{v}$ by \textbf{16.62$\%$} compared by random strategy. Model strategy also increased ROI by \textbf{24.33$\%$} compared by random strategy. 

In daily coupon allocation, model strategy increased $V_{r}$ by \textbf{83.4$\%$} compared by random strategy. Model strategy also increased $O_{v}$ by \textbf{4.55$\%$} compared by random strategy. 

\textbf{Promotion Period.} There are many big online promotions on Taobao every year. We introduces two big promotions which are "Double 9" and "Double 11" from them. These two promotions are big shopping festivals in China, similar as the "Black Friday" in America.

In "Double 9" promotion, the model strategy can effectively optimize the cost and efficiency of subsidies on the premise of ensuring the scale transformation. Compared with the random strategy, the ROI increased by \textbf{241$\%$}.

In "Double 11" promotion, we provide coupons of different categories according to the users' intent predicted by TPG-DNN after people has bought the goods. Model strategy increased $V_{r}$ by \textbf{10.44$\%$} compared by random strategy. 

TPG-DNN can be applied not only in the coupon allocation, but also in the product recommendation. For example, for users who browse a lot but purchase little, their browsing probability are very high and purchase intention are quite low, which indicates that they have purchase demand but may not be faced with the target products. In this situation,we will recommend products for users they maybe interested with our TPG-DNN based on users' historical behaviors, and gradually improve recommendation effects on the basis of user behavior feedback constantly. 

\section{Conclusion}
In this paper, we propose a novel multi-task user intent prediction model. Based on the total probability GRU loss function, the information of user time-series behavior are fully extracted. In addition, we introduce a multi-task weight adjustment mechanism based on the maximum likelihood variance uncertainty, allowing the final loss function to dynamically adjust the weight between different tasks through data variance. According to the test result of our model from Taobao user data, the proposed model perform much better than the existing ones. At present, Our user intent prediction model TAG-DNN has widely supported Taobao in the fields including equity methods, advertising and product recommendation, which will greatly improve the user experience and benefit the GMV promotion on Taobao platform.


\bibliographystyle{ACM-Reference-Format}
\bibliography{sample-base}


\end{document}